\documentclass[letterpaper, 10 pt, conference]{bib/ieeeconf}
\IEEEoverridecommandlockouts                              
\usepackage[normalem]{ulem}	                        
\usepackage[table,usenames,dvipsnames]{xcolor}      
\usepackage{extarrows}                              
\let\labelindent\relax
\usepackage{enumitem}                               

\usepackage{amsmath,amsthm,amssymb,amsfonts,dsfont} 

\makeatletter
\newif\if@restonecol
\newif\foralgo@restonecol
\makeatother

\usepackage[linesnumbered,ruled,vlined]{algorithm2e}
\usepackage{algpseudocode}
\usepackage{amsmath}

\usepackage{graphicx}
\usepackage{subfigure}
\usepackage{tabularx}
\usepackage{amsmath}
\usepackage{multirow,multicol, array}
\usepackage[font={small}]{caption}   
\usepackage[titletoc,title]{appendix}
\usepackage[breaklinks=true, colorlinks, bookmarks=true, citecolor=Black, urlcolor=Violet,linkcolor=Black]{hyperref}

\newcommand{\tabincell}[2]{\begin{tabular}{@{}#1@{}}#2\end{tabular}}  

\newtheorem{proposition}{Proposition}

\newtheorem{definition}{Definition}

\newtheorem*{assumption*}{Assumption}

\newtheorem*{remark*}{Remark}
\newtheorem*{problem*}{Problem}

\title{\LARGE \bf
Model-based Constrained Reinforcement Learning using Generalized Control Barrier Function
}
\author{Haitong Ma$^\ddag$, Jianyu Chen$^\dag$, Shengbo Eben Li$^\ast{}^\ddag$, Ziyu Lin$^\ddag$, Yang Guan$^\ddag$, Yangang Ren$^\ddag$, and Sifa Zheng$^\ddag$
\thanks{$^\ddag$School of Vehicle and Mobility, Tsinghua University. \texttt{Email: \{maht19@mails., lishbo@, linzy17@mails., guany17@mails., ryg18@mails., zsf@\}tsinghua.edu.cn}.}
\thanks{
$^\dag$Institute for Interdisciplinary Information Sciences, Tsinghua University. \texttt{Email: jianyuchen@tsinghua.edu.cn}.}
\thanks{
$^*$All correspondence should be sent to S. Li.}
}

\begin{document}
\maketitle
\begin{abstract}
Model information can be used to predict future trajectories, so it has huge potential to avoid dangerous regions when applying reinforcement learning (RL) on real-world tasks, like autonomous driving. However, existing studies mostly use model-free constrained RL, which causes inevitable constraint violations. This paper proposes a model-based feasibility enhancement technique of constrained RL, which enhances the feasibility of policy using generalized control barrier function (GCBF) defined on the distance to constraint boundary. By using the model information, the policy can be optimized safely without violating actual safety constraints, and the sample efficiency is increased. The infeasibility in solving the constrained policy gradient is handled by an adaptive coefficient mechanism.  We evaluate the proposed method in both simulations and real vehicle experiments in a complex autonomous driving collision avoidance task. The proposed method achieves up to four times fewer constraint violations and converges 3.36 times faster than baseline constrained RL approaches.
\end{abstract}

\section{Introduction}
\label{sec:intro}

Safety is critical when applying reinforcement learning (RL) to real-world tasks \cite{RLBOOK}. For instance, in the field of autonomous vehicle control, the collision must be avoided in case of causing physical harm to humans \cite{amodei2016concrete}. A safety-critical reinforcement learning problem is generally formulated to a constrained reinforcement learning problem, aiming to maximize the reward function while satisfying the safety constraints \cite{Achiam2017a,Chow2018a}. 

Multiple definitions of the cost-based constraints can be integrated with constrained RL. The chance constraint is the most popular choice, where a one-hot design of cost signal is commonly used \cite{Garcia2015a}. Both average cost-based constrained and accumulative cost constraints are considered in different algorithms \cite{Uchibe2007,tessler2018reward}. Value at risk measures risk as the maximum possible cost with a pre-defined confidence level \cite{jorion2007value}. Conditional value at risk (CVaR) is further designed to address those cases whose probability is small, usually used in portfolio optimization \cite{Rockafellar2002}.  Both of them are designed with long-horizon data-driven expectation, which is the inevitable choice for model-free RL. The drawback is that existing model-free RL can only learn a safe policy by inevitably experiencing constraints violations through trial-and-error, which imposes significant safety issues, especially during exploration \cite{Ray2019}.

Some existing constrained RL methods deploy model information to obtain a constraint-satisfying policy. Most existing studies aim to find a constrained optimal policy while adopting constraints on \emph{every time step} in the prediction horizon with model rollout \cite{Duan2019b, memarzadeh2019model}. Some learning-based controllers share the similar idea with multi-step rollout with model and constraints on each time step \cite{koller2018learning}. This design's major problem with pointwise constraints is that the prediction will become inaccurate with the increase of rollout steps. Moreover, a multi-step rollout uses too much sampling information to finish the constrained optimization, and the sampling efficiency is significantly decreased.


In this paper, we propose a model-based constrained reinforcement learning approach with the generalized control barrier function. Intuitively, applying the control barrier function can handle state constraints by penalizing the trends of getting closer to the constraint boundary \cite{RLBOOK}. The proposed GCBF constraints are only considered within one or a few prediction steps, so the sampling efficiency increases, and the issue of prediction inaccuracy is avoided. We apply the approximate Lagrangian  solution technique to compute the constrained policy gradient, and an adaptive mechanism is further added to automatically choose a appropriate parameters to improve the constraint-satisfying performance. The main contribution of this paper is summarized as follows: 

(1) We have fully dug the model's information for constrained RL by penalizing the trends getting closer to the constraint boundary. A constraint-satisfying policy can be learned without violating actual safety constraints. The constraints violations during training are up to 73.83\% lower than baseline constrained RL approaches. 

(2) The constraints formulation has the theoretically smallest required steps in each iteration without learning the cost approximation with proof. The sampling efficiency improves by 3.36 times compared to baseline model-based constrained RL.

The paper is organized as follows. Section II is the preliminaries about the key components of constrained RL and generalized control barrier function. Section III introduces the proposed model-based constrained RL algorithms and the adaptive mechanism to choose GCBF's parameters. Section IV demonstrates the experiment results on the simulation platform and a real autonomous vehicle. Section V concludes the paper.

\section{Preliminaries}
\subsection{Constrained Reinforcement Learning}

Constrained reinforcement learning (RL) indicates the general problem of training an RL agent with constraints, usually with the intention of satisfying constraints throughout exploration in training and at test time.

\begin{equation}
\pi^{*}=\operatorname{argmin}_{\pi \in \Pi_{C}} J_{r}(\pi) \label{eq1} 
\end{equation}
where $J_r (\pi)$ is the expected return. The feasible policy set $\Pi_C$ is determined by inequality constraints, mostly in a cost-based formulation:
\begin{equation}
\Pi_{\mathrm{C}}=\left\{\pi: J_{C_{i}}(\pi) \leq d_{i}\right\}
\end{equation}
where $i=1,2,…,k$ is the constraint index. Each $J_{C_i}$ is the expected cost, and $d_i$ is a pre-defined threshold. Recently, numerous efforts to improve constrained RL are based on the actor-critic architecture integrated with the “constrained policy optimization” technique. The actor update progress is modified to find a constraint-satisfying policy, and the critic update is the same as existing state-value RL algorithms like trust-region policy optimization (TRPO) \cite{schulman2015trust,Lin2021Solving}. 

\subsection{Formulations of Inequality Constraints}

Constraint formulations directly affect the safety performance,  which is critical in constrained RL. An early CAC-like algorithm, i.e., the policy gradient projection (PGP), whose constraints formulation is based on average cost\cite{Uchibe2007}:
\begin{equation}
\lim _{T \rightarrow \infty}\left[\mathbb{E}_{s \sim d(s), a \sim \pi_{k}}\left(\frac{1}{T} \sum_{t=1}^{T} r_{c_{i}}\right)\right] \leq d_{i}
\end{equation}
where $r_{C_i}$ is the corresponding constraint cost in a one-hot formulation, where a constraint-violation action gets a cost of one. The average reward design is not able to handle the unsafe action with a low probability. Chow et, al. (2015) instead adopt constraints on conditional value at risk (CVaR) \cite{Chow2018a}:
\begin{equation}
\min _{v \in \mathbb{R}}\left\{v+\frac{1}{1-\zeta} \mathbb{E}_{s \sim d(s), a \sim \pi_{k}}\left[\left(r_{C_{i}}-v\right)^{+}\right]\right\} \leq d_{i}
\end{equation}
The confidential level $\zeta$ is a pre-defined hyperparameter, $v$ is a balance coefficient between reward and cost. CVaR is about to address the actions in low probability but severer consequences. However, the balancing parameters design still accepts some constraints violations, which is not appropriate for the safety-critical problems. Later, the famous constrained policy optimization (CPO) algorithm is proposed, which firstly claims to guarantee safe exploration \cite{Achiam2017a}. The constraints formulation is the accumulative constraint costs with a trust-region constraint to bound the constraint performance:
\begin{equation}
\overline{D_{p}}\left(\pi_{k}, \pi_{k+1}\right) \approx \frac{1}{2} \Delta \theta^{T} H \Delta \theta<\delta
\end{equation}
where $\overline{D_{p}}$ is a distance measurement. In practice, $\overline{D_{p}}$ is replaced with the KL divergence with second-order Taylor approximation, $H$ is the Fisher information matrix. CPO is regarded as a commonly used baseline of model-free safe RL. 
 
 A typical model-based policy optimization (MBPO) for constrained RL is proposed by Duan et, al. (2019). It adopts multi-step rollout to confine policy update, where the constraints are separately posed on \emph{each rollout step} \cite{Duan2019b}:
\begin{equation}
J_{C_{i}}\left(\pi_{k}\right)=\mathbb{E}_{a \sim \pi_{k}}\left\{r_{C_{i}}\left(s_{t+i}, a\right)\right\} \leq d_{i}
\end{equation}
where $i \in {1,2,…N}$, and $\forall s_t$ in the safe state set. Each policy update needs an N-steps model rollout. The comparison between four typical algorithms is listed in TABLE. \ref{algos}. CPO and MBPO are chosen as the baselines of our proposed algorithms.
\begin{table}[h]
\centering
  \caption{Constraint Formulations of Typical CAC algorithms} \label{algos}
\begin{tabular}{cc}
\hline
Algorithms & Constraints formulation                          \\ \hline
PGP        & Average cost constraint                          \\
PDO        & Conditional value at risk                        \\
CPO        & Accumualtive cost constraints \& trust region    \\
MBPO       & Model-based statewise constraint \& trust region \\ \hline
\end{tabular}
\end{table}

In summary, the cost-based constraints usually adopted in model-free constrained RL are learned with experiencing the constraint violations,  which causes significant safety issues. The model-based approaches pose constraints based on the multi-step rollout, which causes problems with the low sampling efficiency and inaccuracy prediction in the future rollout steps. All of these issues block the performance improvement of existing constrained RL.

\subsection{Generalized Control Barrier Function}
Aforementioned methods all directly adopts constraints formulation with $h(\cdot)\leq 0$. On the contrary, control barrier function (CBF) adopts a more concise formulation. Control barrier function is proposed to address safety with dynamic systems, also called the safety barrier certificate \cite{prajna2006barrier, Agrawal2017a}. We define a safe state set concerning real-world safety requirements:
\begin{equation}
\mathcal{C}=\{s \mid h(s) \leq 0\}
\end{equation}

Consider a general discrete-time dynamical system:
\begin{equation}
\label{dynamics}
s_{t+1}=f\left(s_{t}, a_{t}\right)
\end{equation}

\begin{definition}[Control barrier function]
The discrete-time control barrier function (CBF) for a constraint $h(s_t )\leq0$ is 
\begin{equation}
h\left(s_{t+1}\right) \leq(1-\alpha) h\left(s_{t}\right)
\end{equation}
where $\alpha$ is the conservativeness coefficient. 
\end{definition}

\begin{figure}[ht]
\centering
\subfigure[Traditional pointwise constraints]{\includegraphics[width=0.48\linewidth]{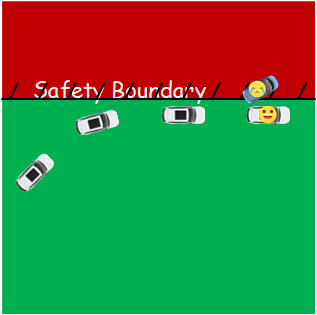}}
\subfigure[Control barrier function.]{\includegraphics[width=0.48\linewidth]{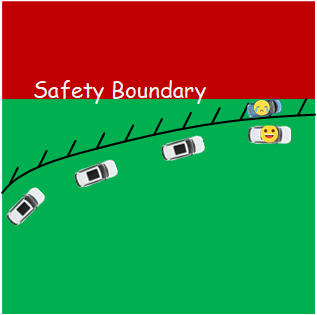}}
\caption{Intuitive explanation of control barrier functions.}
\label{fig:cbf}
\end{figure}

For a constrained set $\mathcal{C}$ with a CBF constraint is satisfied for all states, the set can be guaranteed safe with respect to the system (\ref{dynamics}). Intuitively, control barrier function can be explained by confining the trend of getting closer to the constraint boundary shown in Fig. \ref{fig:cbf}. A larger $\alpha$ indicates that the constraints are less conservative. 

A major drawback of the original formulation is that it cannot be applied on high relative-degree dynamic systems \cite{Nguyen2016,Ames2019a}. The relative-degree is defined as which order derivative of constraints is relevant with the control input, i.e.,
\begin{definition}[High relative-degree constraints]
\label{def:degree}
The constraint has relative-degree $m$ with respect to control input if
\begin{equation}
\frac{d h\left(s_{t+m}\right)}{d s_{t+m}}\frac{d f(s_{t+i-1}, a_{t+i-1})}{d a_{t}}=0
\end{equation}
for $\forall i \in\{0,1, \ldots m-1\}, \forall s_t \in \mathbb{R}^{n}$, with respect to system  \ref{dynamics}, $m\in\{2,3,…n\}.$ If the above equality does not hold, the constraint has relative-degree 1. 
\end{definition}
In our previous work, we propose the generalized control barrier function to handle high relative-degree constraints is to pose constraints on the nonadjacent steps for a constraint function with arbitrary relative-degree $m$.

\begin{definition}[Generalized Control Barrier Function]
For a constraint with relative degree $m$, the generalized control barrier function is 
\begin{equation}
h\left(s_{t+m}\right) \leq(1-\alpha)^{m} h\left(s_{t}\right), \forall k \in \mathbb{Z}_{+}
\end{equation}
\end{definition}
The intuitive explanation is that the high-order derivatives are “flatten” on the time axis. In order to track the input, the constraint is posed between two nonadjacent steps. Details about discrete-time control barrier function are provided in our previous work \cite{ma2021feasibility}.

\section{Algorithm Details}
This section introduces how to confine policy updates by GCBF, including the problem formulation, the approximate update rules, and an adaptive conservativeness mechanism to correct the parameters in control barrier function.
\subsection{Model-based Policy Optimization with GCBF}
\subsubsection{Problem formulation}

A reinforcement learning algorithm is to optimize the expected returns. The critic and actor need to be updated during the policy optimization.  Defining the return as $\sum_{j=t}^{t+m} \gamma^{j-t} r\left(s_{j}, \pi\left(s_{j} ; \theta\right)\right)+\gamma^{m} V\left(s_{t+m+1} ; w\right)$, the actor update stage is a constrained optimization with the GCBF constraints, where the optimization problem is:
\begin{equation}
\begin{aligned}
\min _{\Delta \theta}\ J_r(\theta)&=\mathbb{E}_{s \sim \mathcal{C}, a \sim \pi(\theta)}\{G\} \\
\text{s.t.}\ J_{C_{i}}(\theta)&=\mathbb{E}_{a \sim \pi(\theta)}\left[h_i\left(s_{t+m}\right)\right] \\
&\leq(1-\alpha)^{m} h_i\left(s_{t}\right)
\end{aligned}
\label{eq:problem}
\end{equation}
Note that the $J_{C_i }(\theta)$ is calculated by $m$-steps rollout with models. The original MBPO uses a multi-step rollout, for example, 10-steps setting in the original paper, as a constrained prediction horizon, while we only need $m$-steps information to finish a policy update. The following section will demonstrate that the efficiency improvement. 
\begin{proposition}[Least Required Sampling Steps]
\label{prop:leaststep}
For a constraint with relative-degree $m$, the model-based constrained policy optimization should rollout at least $m$ steps.
\end{proposition}
The proof is provided in Appendix. The critic update rule is similar to the unconstrained version, where the critic loss is defined as
\begin{equation}
L(w)=\mathbb{E}_{s_t \sim \mathcal{C}}\left\{\frac{1}{2}\left(G-V\left(s_{t} ; w\right)\right)^{2}\right\}
\end{equation}
and the gradient of critic is 
\begin{equation}
\frac{d L}{d w}=\mathbb{E}_{s_t \sim \mathcal{C}}\left\{\left(G-V\left(s_{t} ; w\right)\right) \frac{d V\left(s_{t} ; w\right)}{d w}\right\}
\label{criticgrad}
\end{equation}

The critic update has not changed compared to the unconstrained version. 
\subsubsection{Approximate Solution for Constrained Policy Gradient}

The gradient $\Delta \theta$ to update actor must satisfy (\ref{eq:problem}). We implement the approximate solution technique by linearized objective and constraints added with a distance constraint.
\begin{equation}
\begin{aligned}
\min _{\Delta \theta}\  &g^{T} \Delta \theta\\
\text { s.t. }& z+C^{T} \Delta \theta \leq 0 \\
&\overline{D_{p}}\left(\theta ; \theta_{k}\right) \approx \frac{1}{2} \Delta \theta^{T} H \Delta \theta \leq \delta
\end{aligned}
\label{lin_lag}
\end{equation}
where $g=\frac{d J}{d \theta} /\left\|\frac{d J}{d \theta}\right\|^{2}, \cdot z_{i}=\left(\left.J_{C_{i}}\right|_{\theta_{k}}-(1-\lambda)^{m} h\left(s_{k}\right)\right)$, $C_{i}=\frac{d J_{C_{i}}}{d \theta} /\left\|\frac{d J_{C_{i}}}{d \theta}\right\|^{2}$. With $C \doteq\left[c_{1}, c_{2}, \ldots, c_{M}\right]$ and $z \doteq\left[z_{1}, z_{2}, \ldots, z_{M}\right]$, the the analytical solution of (\ref{lin_lag}) can be analytically solved by Lagrange multiplier method. The Lagrange function are
\begin{equation}
\begin{aligned}
L(\Delta \theta, \lambda, v)
=&g^{T} \theta +\lambda\left(\frac{1}{2} \Delta \theta^{T} H \Delta \theta-\delta\right)\\&+v(z
\left.+C^{T} \Delta \theta\right) 
\end{aligned}
\end{equation}
where $\lambda, \nu$ is the dual variable.
The analytical optimal solution is 
\begin{equation}
\Delta \theta^{*} =\frac{H^{-1}\left(g+C v^{*}\right)}{\lambda^{*}}
\label{feasibleupdate}
\end{equation}
where $\lambda^*, \nu^*$ is the optimal dual solution obtained by analytical solution (single-dimension constraint) or solvers (multi-dimension constraints). If the problem does not have a feasible solution, the policy update rule changes to a retrieval mechanism:
\begin{equation}
\theta_{k+1}=\theta_{k}-\sqrt{\frac{2 \delta}{b^{T} H^{-1} b}} H^{-1} b
\label{infeasiupdate}
\end{equation}
 The pseudocode is shown in Algorithm 1.

\begin{algorithm}
    \label{algo:ori}
    \caption{GCBF-MBPO}
    
    \LinesNumbered
    
    \KwIn{Feasible policy $\pi(\theta_0 )$, constraint relative degree $m$, conservativeness coefficient $\alpha$}
    
    \For{$k=1,2,\dots$}
    {
    Sample a set of trajectories $\mathcal{D}=\{\tau\} \sim \pi_{k}=\pi\left(\theta_{k}\right)$
    
    From samples predicts $g, b, H, c$
    
    \If{ appximate update is feasible }
    {
        Solve dual problem and update theta with (\ref{feasibleupdate})
    }
    \Else{Compute recovery policy with (\ref{infeasiupdate})}
    
    Update critic with (\ref{criticgrad})
    }
    
\end{algorithm}
\subsection{Adaptive Conservativeness Mechanism}

Intuitively, a more conservative choice of $\alpha$ in CBF may lead to more retrieval updates and affects the constraint-satisfying performance. To find a proper conservativeness coefficient, we propose an adaptive updating rule of $\alpha$, which adjusts the value according to the severity of violations of the GCBF constraint. We predict the constraints violation $\xi$ from the trajectory $\mathcal{T}$:
\begin{equation}
\xi=\mathbb{E}_{\mathcal{T}} \sum_{i}\left[J_{C_{i}}(\pi)-d_{i}\right]^{+}
\end{equation}
If the constraints violation exceeds a pre-defined threshold, the conservativeness coefficient is adjusted to releases the constraints. We name the modified version with adaptive conservativeness coefficient as adaptive $\alpha$ GCBF-MBPO, shown in Algorithm 2.

\begin{algorithm}
    \label{algo:modi}
    \caption{Adaptive $\alpha$ GCBF-MBPO}
    
    \LinesNumbered
    
    \KwIn{Feasible policy $\pi(\theta_0 )$, constraint relative degree $m$, conservativeness coefficient $\alpha$, violation tolerance $\xi_c$}
    
    \For{$k=1,2,\dots$}
    {
    Sample a set of trajectories $\mathcal{D}=\{\tau\} \sim \pi_{k}=\pi\left(\theta_{k}\right)$
    
    From samples predicts $g, b, H, c, \xi$
    
    \If{ appximate update is feasible }
    {
        Solve dual problem and update theta with (\ref{feasibleupdate})
    }
    \Else{Compute recovery policy with (\ref{infeasiupdate})}
    
    Update critic with (\ref{criticgrad})
    
    \If{$\xi>\xi_c$}
    {
    $\alpha \leftarrow \alpha+\beta \xi$
    }
    }
    
\end{algorithm}
\section{Experimental Results}
\label{sec:exp}
\begin{figure}[h]
\centering
\includegraphics[width=0.9\linewidth]{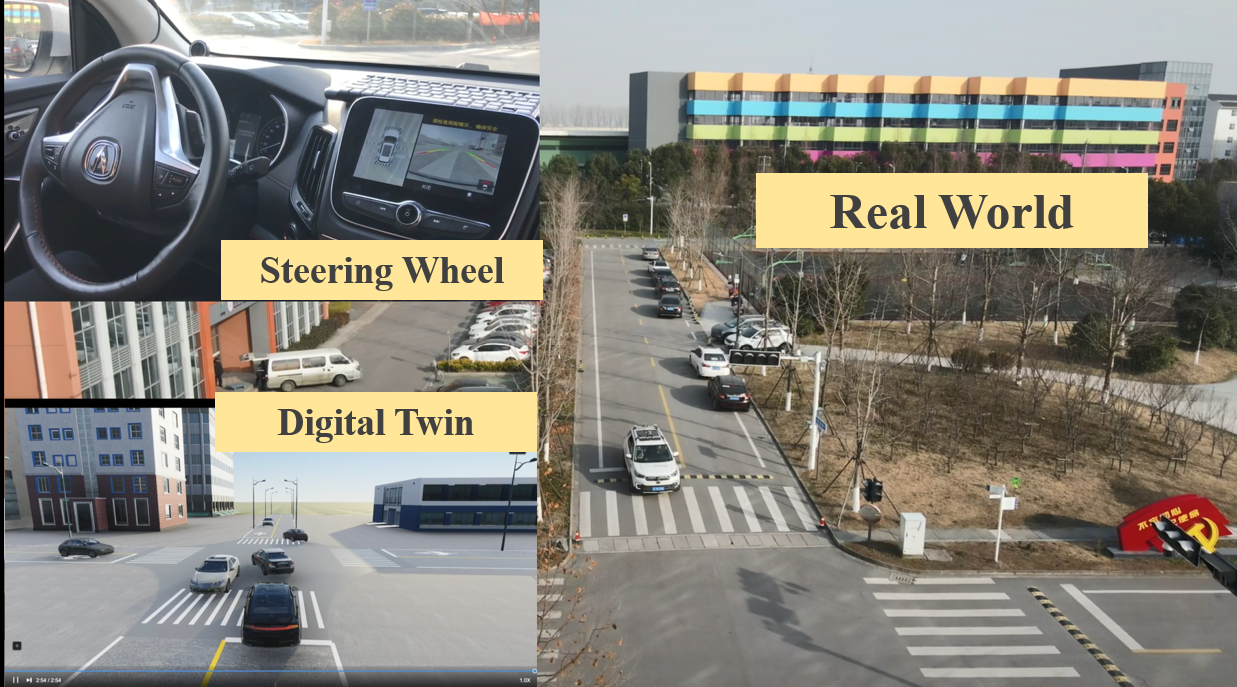}
\caption{The autonomous vehicle collision avoidance with a digital twin system.}
\label{fig:autoexp}
\end{figure}

Autonomous driving is a complex safety-critical sequential decision-making problem with multi-objective orientation, which poses great challenges to decision and control systems \cite{Li2011}\cite{Li2015}. The intersection is a complex scenario for autonomous driving, where collision avoidance is the major safety concern \cite{Guan2020Centralized} \cite{Ren2020ITSC}.  This section evaluates the proposed algorithms on a large-scale autonomous driving task in a two-way six-lane intersection to show the constraints violations reduction and efficiency improvements. We also apply our proposed algorithm to a real autonomous vehicle to verify the collision avoidance ability. The surrounding vehicles are generated virtually by a digital twin system for the safety consideration shown in Fig. \ref{fig:autoexp}.

\subsection{Experiment 1: Simulation}
\subsubsection{Problem Description}
\begin{figure}[h]
\centering
\includegraphics[width=0.8\linewidth]{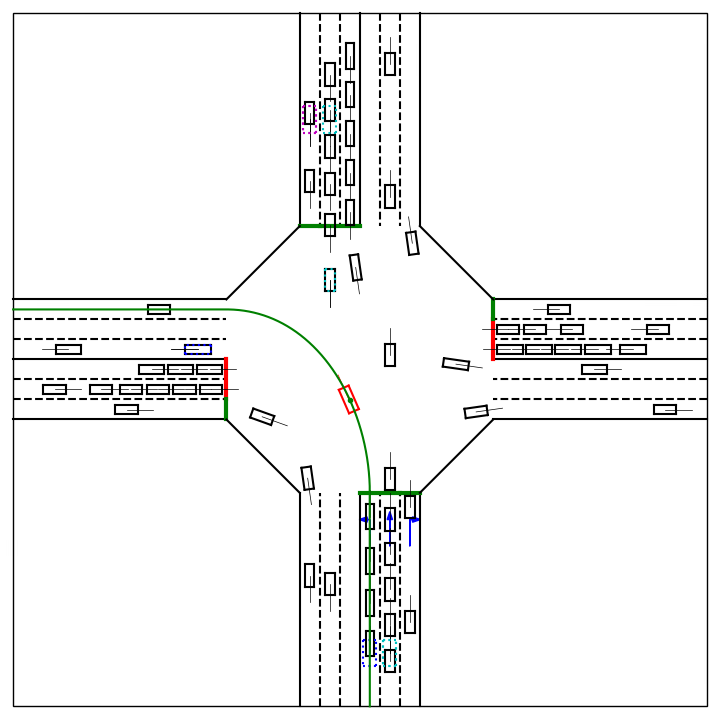}
\caption{The intersection for autonomous driving control task. We wrap the scenario as a safety-gym third party environment, the code repo is on \url{https://github.com/mahaitongdae/safe_exp_env}.}
\label{demo}
\end{figure}

The autonomous driving task requires the agent to track the pre-defined reference path to pass the intersection without colliding into other vehicles or road margins.  The intersection is demonstrated in Fig. \ref{demo}, and the random traffic flow is generated by SUMO. 

The states include both states of ego vehicle, tracking error, and surrounding vehicles. All surroundings are filtered to 8 involved vehicles according to the distance to ego vehicle and each vehicle’s goal lane. If the number of involved vehicles is less than 8, certain virtual vehicles are augmented with a distant location. The dimension of state space sums up to be 41, and the action includes desired acceleration and steering angle of the ego vehicle. Details are listed in TABLE \ref{tab:sa}.

\begin{table}[]
\caption{State and Control Input}
\label{tab:sa}
\begin{tabular}{lccc}
\hline 
Ego vehicle state                       & Speed         &$(v_x,v_y)$  & {[}m/s{]}   \\
                                        & Yaw rate                   &$r_y$  & {[}rad/s{]} \\
                                        & Position      &$(x,y)$  & {[}m{]}     \\
                                        & Heading angle              &$\psi$ & {[}rad{]}   \\
                                        \hline 
Tracking states                         & position error     &$(\Delta x,\Delta y)$  & {[}m{]}     \\
                                        & Heading angle error        &$\Delta\psi$  & {[}rad{]}   \\
                                        \hline 
Surrounding vehicle states & Position         &$(x_j, y_j)$   & {[}m{]}     \\
                                        & Velocity                   &$v_j$  & {[}m/s{]}   \\
                                        & Heading angle              &$\psi_j$  & {[}rad{]}   \\
                                        \hline 
Input                                   & Steering anlge             &$\delta$  & {[}rad{]}   \\
                                        & Acceleration               &$a_{Acc}$  & {[}m/s2{]}\\
                                        \hline
\end{tabular}
\end{table}
The reward function is formulated to track a static trajectory randomly selected to reach each destination lane:
\begin{equation}
\begin{array}{c}
r(s, a)=0.05\left(v-v_{\text {target }}\right)^{2}+0.8 \Delta y^{2}+30 \Delta \phi^{2} \\
+0.02 r_{y}^{2}+5 \delta^{2}+0.05 a_{\text {Acc }}^{2}
\end{array}
\end{equation}

\begin{figure}[h]
\centering
\includegraphics[width=0.8\linewidth]{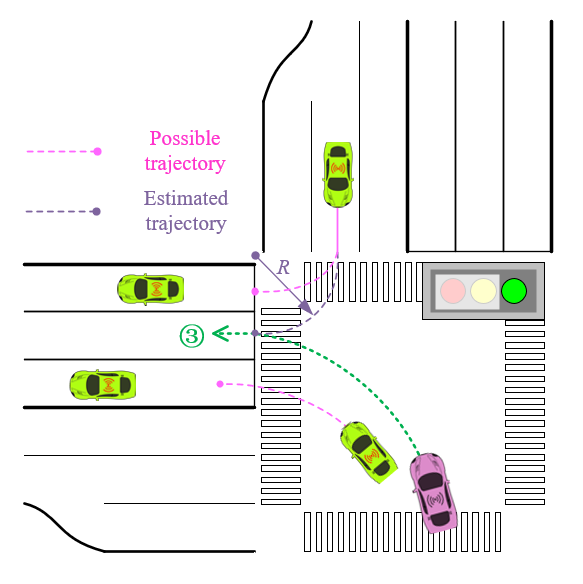}
\caption{Predicting surrounding vehicles.}
\label{fig:predict}
\end{figure}

The model of ego vehicle uses a numerically stable dynamic bicycle model \cite{ge2020numerically}.  As for the surrounding vehicles, a simple kinematics model with the uniform recurrence assumption is adopted. The target for each surrounding vehicle can be obtained from SUMO, which tells whether a vehicle prepares to go straight, turn left, or right. The states for position information are predicted with uniform recurrence driven by current speed, and the yaw angle is predicted by the constant-speed rotation, i.e.,
\begin{equation}
\begin{aligned}{r}
x_{i}^{\prime}&=x_{i}+v_{i} \cos \left(\phi_{i}\right) T \\
y_{i}^{\prime}&=y_{i}+v_{i} \sin \left(\phi_{i}\right) T \\
\phi_{i}^{\prime}&=\left\{\begin{array}{ll}
\phi_{i} & \text { if going straight } \\
\phi_{i}+\frac{v_{i}}{R^{*}} T & \text { if turning }
\end{array}\right.
\end{aligned}
\end{equation}
where $R^*$ is an estimated radius depending on the intersection's size demonstrated in Fig. \ref{fig:predict}. For instance, in the simulation scenario, the intersection's size is 50 m, and the turning radius of the right turn is 20 m, while the left turn is 30 m. Both ego and surroundings model is not perfect, but the results section will show a considerable reduction of constraints violation.

\begin{figure}[h]
\centering
\includegraphics[width=0.8\linewidth]{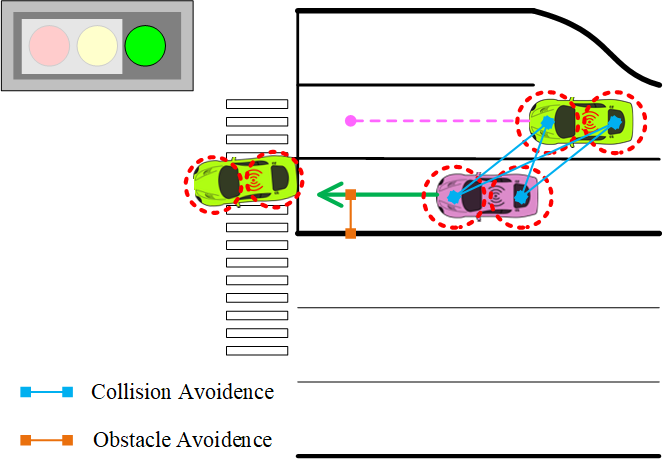}
\caption{Demonstration of state constraints.}
\label{fig:cstr}
\end{figure}

The safety constraints include collision avoidance and road margin. A two-circles safe distance constraint is implemented between the ego vehicle and each vehicle:
\begin{equation}
\begin{aligned}
\left(x^{\#}-x_{j}^{*}\right)^{2}+\left(y^{\#}-y_{j}^{*}\right)^{2} &\geq d_{\text {safe }}^{2} \\
\left(x^{\#}-x_{\text {road }}\right)^{2}+\left(y^{\#}-y_{\text {road }}\right)^{2} &\geq d_{r_{\text {safe }}}^{2}
\end{aligned}
\end{equation}
where $(x^*,y^*)$ is the center of circles, and the subscripts $j\in{1,2,…8}$ represents the index of surrounding vehicles. The up-scripts  $\#,*\in\{f,r\}$ represents the front or rear safety circle as shown in Fig. \ref{fig:cstr}. The road margin is also considered similar to the two-circles safety distance constraints, where the nearest point to the road margin is represented by $(x_{road},y_{road})$. 

\subsubsection{Training Results}
We compare our adaptive $\alpha$ GCBF-MBPO (Ada-GCBF-MBPO) and the original version (GCBF-MBPO) with model-based policy optimization with original constraints (MBPO) and model-free constrained policy optimization (CPO). The number of environment interactions is limited to 2 million. The hyperparameters are listed in TABLE \ref{table:params}. 

\begin{table}[h]
\centering
\caption{Algorithms Hyperparameters}
\label{table:params}
\begin{tabular}{lc}
\hline
Algorithms                       & Value                        \\ \hline
\emph{shared}                           &                              \\
\quad Optimizer                        & \tabincell{c}{Conjugate gradient\\optimizer}  \\
\quad Damping coefficient              & 0.1                          \\
\quad Backtracking coefficient         & 0.8                          \\
\quad Max backtracking iterations      & 10                           \\
\quad Approximation function           & Multi-layer perceptron       \\
\quad Number of hidden layers          & 2                            \\
\quad Number of hidden units per layer & 256                          \\
\quad Nonlinearity of hidder layer     & ELU                          \\
\quad Nonlinearity of output layer     & tanh                         \\
\quad Critic learning rate             & \tabincell{c}{Linear Annealing\\8e-5$\to$ 8e-6}            \\
\quad Discounted factor                & 0.99                         \\ \hline
\emph{GCBF-MBPO}                        &                              \\
\quad Conservativeness coefficient     & 0.3                          \\
\quad Constraints relative-degree      & 3                            \\ \hline
\emph{Adaptive $\alpha$ GCBF-MBPO}                              &                              \\
\quad Initial $\alpha$                         & 0.1                          \\
\quad Violation tolerance              & 0.3                          \\
\quad $\alpha$ learning rate                    &             1e-3     \\ \hline
\emph{MBPO}                                 &                              \\
\quad Constrained rollout steps        & 10      \\
\hline
\end{tabular}
\end{table}

The average episode returns and episode constraints violation distance are chosen to evaluate the performance of algorithms. The average episode returns are defined with the expectation of episode returns and the feasibility performance, i.e., the constraints violation distance is calculated by for a trajectory $\mathcal{T}$: 
\begin{equation}
\mathbb{E}_{\mathcal{T}} \sum_{j, \#, *}\left[d_{\text {safe }}^{2}-\left(x^{\#}-x_{j}^{*}\right)^{2}+\left(y^{\#}-y_{j}^{*}\right)^{2}\right]^{+}
\end{equation}
where $[\cdot]^+$ represents the positive part, i.e., the violation level of the inequality constraints, the smaller constraints violation distance is, the better feasibility performance algorithm shows. The performance during the training procedure is shown in Fig. 4 and Fig. 5. Results show that the original version of GCBF-MBPO has already decreased the constraints violation by a considerable decent. The performance is not that stable, where lower constraints violations exist in the middle stages of training. The adaptive $\alpha$ mechanism can automatically handle the performance-feasibility balance and keep lower constraint violations throughout the training process.

\begin{figure}[h]
\centering
\includegraphics[width=1\linewidth]{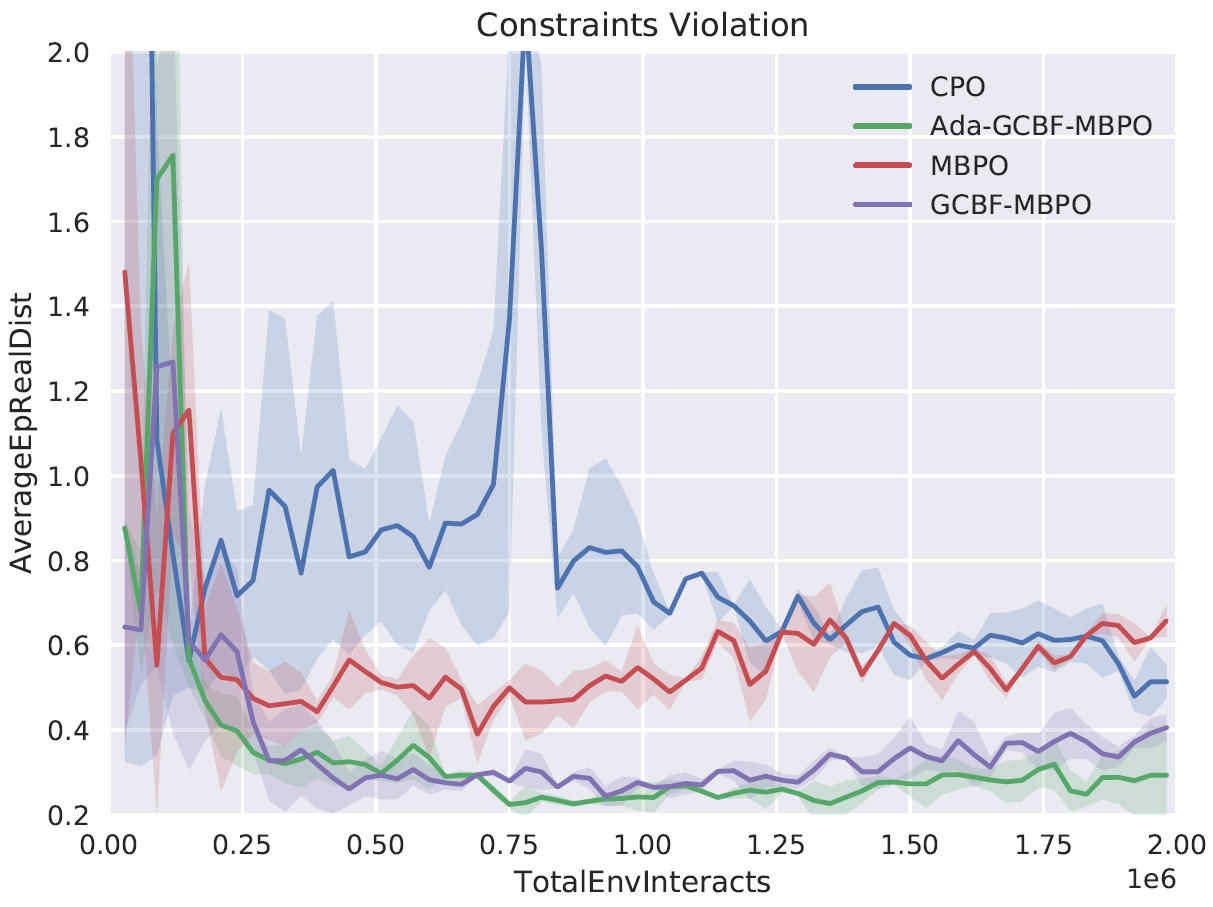}
\caption{Average episode constraints violation distance with different algorithms.}
\label{fig:cstrp}
\end{figure}
\begin{figure}[h]
\centering
\includegraphics[width=1\linewidth]{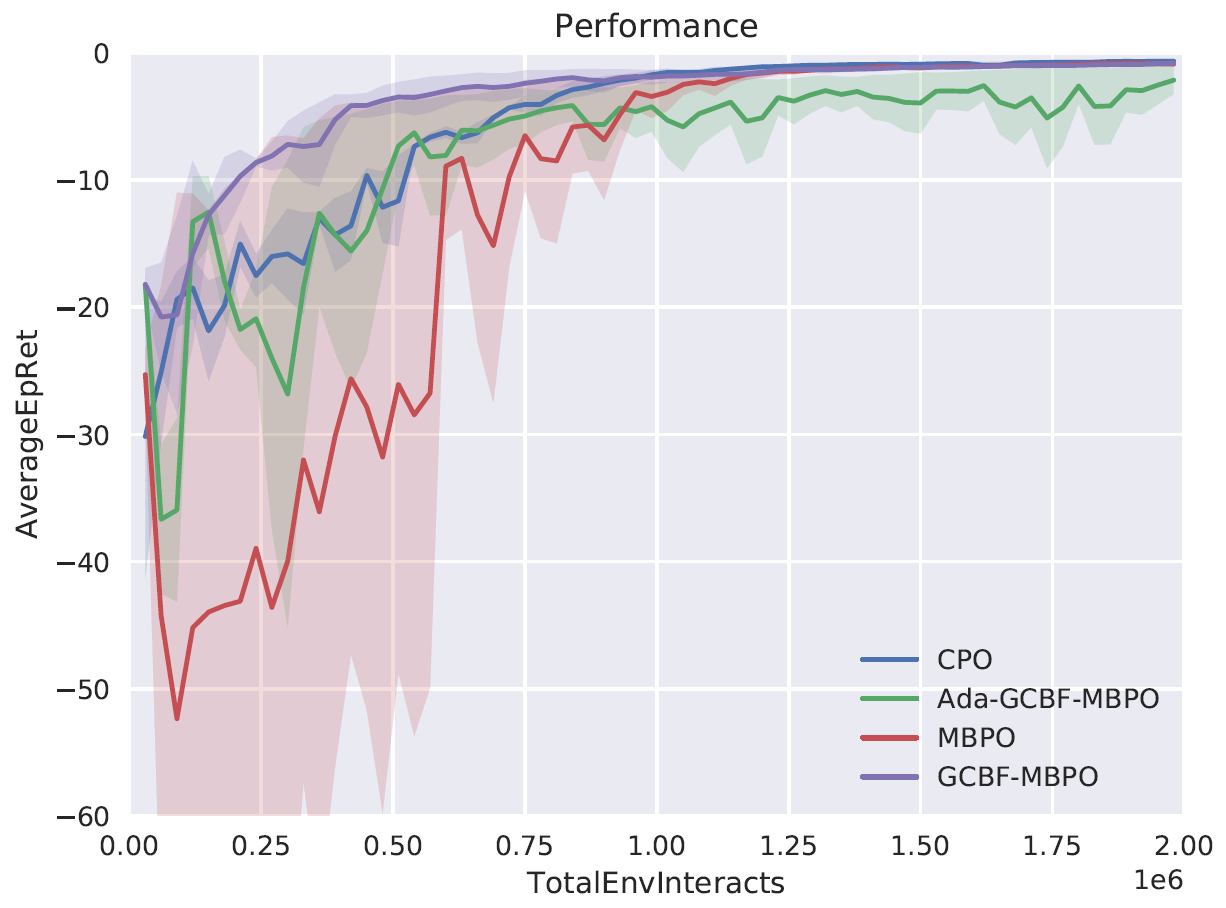}
\caption{Average episode return with different algorithms.}
\label{fig:p}
\end{figure}

The exact numbers of performance and constraints violation distance are shown in TABLE \ref{table:perf}, which demonstrates that GCBF-MBPO can reduce the constraints violation during training from 24.14\% to 73.83\%, while the performance only changes in a reasonable range. Furthermore, it is easy to see the two GCBF-MBPO converges much faster than MBPO algorithms with respect to total environment interactions. We take the total environment interactions when the average episode return reaches several thresholds (-20, -10, -5). The average environment interactions of two GCBF-MBPO are 3.36 times faster than MBPO.

\begin{table}[h]
\caption{Algorithms Performance}
\label{table:perf}
\centering
\begin{tabular}{ccc}
\hline
Algorithms & \tabincell{c}{Average Episode\\Constraints violation} & \tabincell{c}{Average Episode\\Return} \\ \hline
Adaptive  $\alpha$ GCBF-MBPO                      & 0.169                                 & -1.052                 \\
GCBF-MBPO                       & 0.374                                 & -0.769                 \\
MBPO                            & 0.493                                 & -0.785                 \\
CPO                             & 0.646                                 & -0.735                 \\ \hline
\end{tabular}
\end{table}

\subsection{Experiment 2: Autonomous Vehicle}

 Limited by the autonomous driving test regulations, we instead choose a two-lane intersection to demonstrate the vehicle experiment. 
 \begin{figure}[h]
\centering
\includegraphics[width=0.8\linewidth]{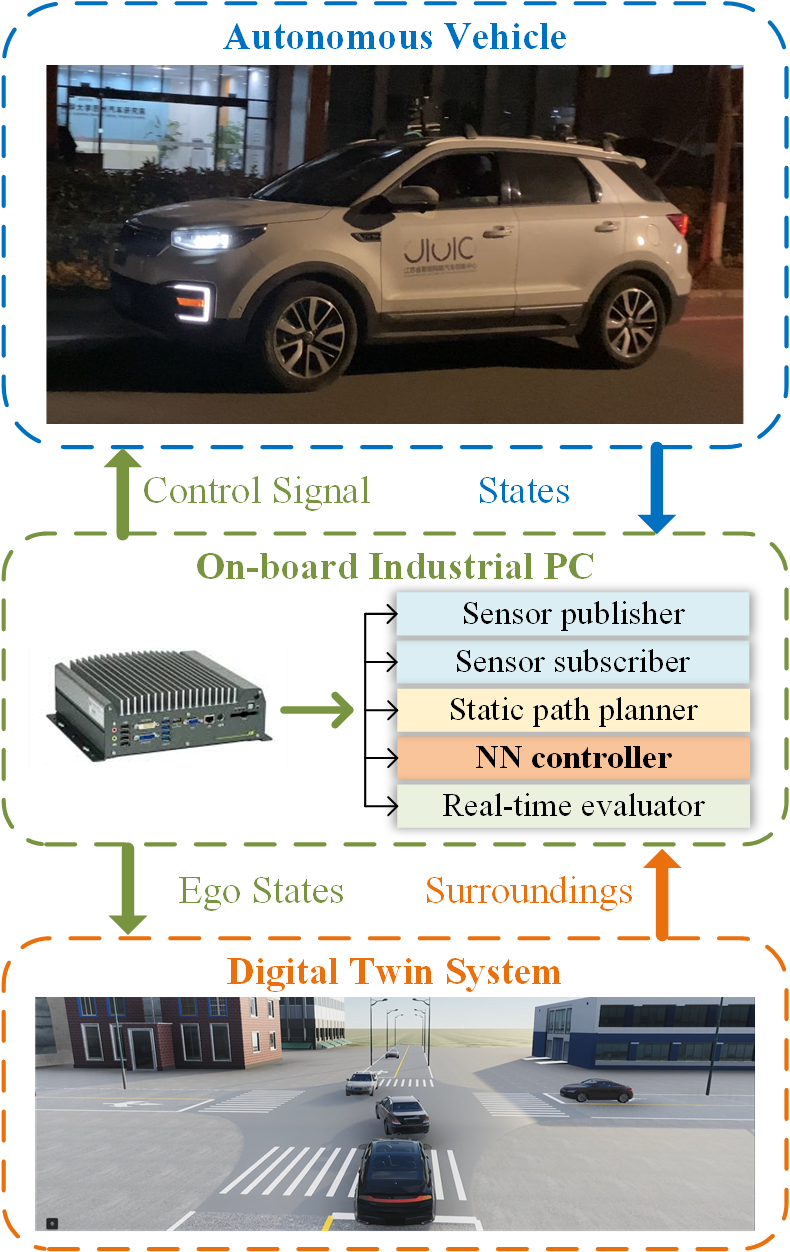}
\caption{Hardware and software architecture of autonomous vehicles.}
\label{fig:arch}
\end{figure}
 \begin{figure*}[t]
\centering
\subfigure[Typical cases.]{\includegraphics[width=0.45\linewidth]{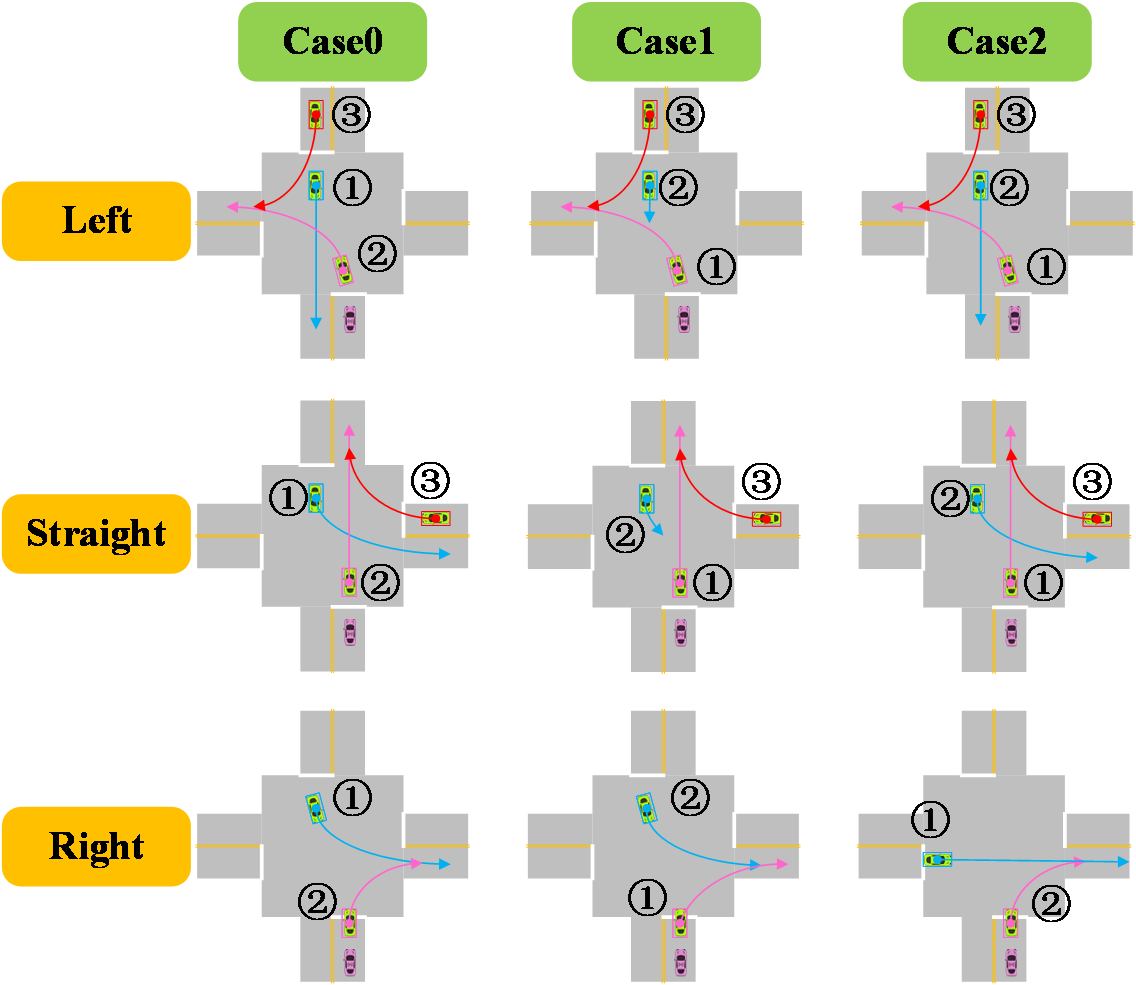}}
\qquad
\subfigure[Bird views.]{\includegraphics[width=0.45\linewidth]{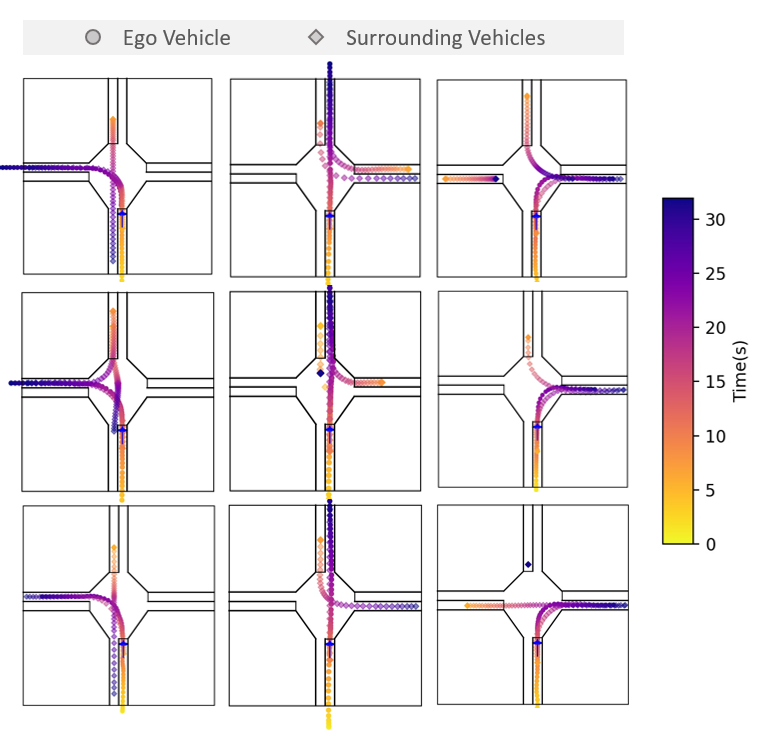}}
\caption{Autonomous vehicle experiments. A short movie is provided to demonstrate the avoidance behaviors on \url{https://youtu.be/WCL2kei0Va0}  or \url{https://b23.tv/k22nVZ}. We select 3 typical cases to demonstrates the autonomous driving vehicle is able to learn avoiding collision by pulling up, decelerating, accelerating and turning. Three perspectives are recorded including autonomous vehicle, steering wheel and digital twin system.}
\label{fig:bdview}
\end{figure*}
 \subsubsection{Hardware and Software Architectures} The autonomous vehicle is a Chang-An CS55 equipped with an on-board industrial PC as the controller. A digital twin-system is adopted to simulate surrounding virtual vehicles. The information of the ego vehicle is also sent back to project the real vehicle in the virtual environment.  The details of hardware and software architecture are shown in Fig. \ref{fig:arch}. Parallel structure is designed in the on-board PC, including neural-network-based controller and planner. 
\subsubsection{Experiment Results}
We select nine typical cases of surrounding vehicles with 3 cases for each destination to test the collision avoidance performance, shown in Fig. \ref{fig:bdview}(a). We demonstrate the experiment from three perspectives, including real-world and virtual environments, as shown in Fig \ref{fig:autoexp}. The arrows represent the surrounding vehicle trajectories, and the indexes are the order to pass the intersection. The results are demonstrated in Fig. \ref{fig:bdview}(b), which includes the time sequences to show the collision avoidance behaviors. Results show that trained policy learns multiple approaches for avoiding collision, including deceleration, accelerating, pulling up and wait, deviating the reference to bypass the vehicles, listed in TABLE \ref{table:behave}.
\begin{table}[h]
\caption{Collision Avoidance Behaviors}
\label{table:behave}
\begin{tabular}{ccccc}
\hline
\tabincell{c}{Destinations}& Decelerating & Pulling up & Accelerating & Turning  \\ \hline
Left                          & case 1,2     & case 0     & -            & case1    \\
Straight                      & case 0,1,2   & case 2     & -            & case 0,1 \\
Right                         & case 2       & -          & case 0       & case 1   \\ \hline
\end{tabular}
\end{table}

\section{Conclusion}
\label{sec: con}
In this paper, we proposed a model-based constrained policy optimization technique with the generalized control barrier function. The model information was utilized to penalize actions that drive agents closer to the constraint boundary. By the proposed approach, learning a constraint-satisfying policy did not need to violate real-world safety constraints. Compared to the baseline model-based constrained policy optimization technique, the efficiency was improved to the maximum with a proof for reducing each policy update's required sampling steps. We further designed an adaptive conservativeness coefficient to handle the infeasibility issue. We evaluate the proposed framework on a collision avoidance task on simulation scenarios and a real autonomous vehicle. Compared to baseline constrained RL, the constraints violation during training decreased by up to 73.83\%, and the efficiency increased 3.36 times. We verified the algorithm functions on the actual autonomous driving vehicles, and the results showed that the policy learned multiple modals of behaviors to avoid collisions.

Although the proposed approach can improve constraint-satisfying performance by model information, the constraints violations still happened due to the approximate solution technique. In the future, we will develop proper solution techniques like augmented Lagrangian to improve the feasibility performance further.

\section*{Acknowledgment}
This study is supported by National Key R\&D Program of China with 2018YFB1600600. This study is supported in part by the Natural Science Foundation of Jiangsu Province under Contract BK20200271 and Suzhou Science and Technology Project under Contract SYG202014. This study is also supported by Tsinghua University-Toyota Joint Research Center for AI Technology of Automated Vehicle. The authors would like to thank Mr. Wei Xu and Prof. Bo Cheng for their valuable suggestions in the autonomous vehicle experiments.
\section*{Appendix}
\begin{proof}[Proof of Prop.~\ref{prop:leaststep}]
Assume a constraint $J_{C_i} (\theta)$ is defined with an expectation of $q$-steps rollout smaller than $m$, the gradient of constraints with respect to actor parameters are
\begin{equation}
\begin{aligned}
\frac{\mathrm{d} J_{C_i}}{\mathrm{~d} \theta}=&\underset{s_{t} \sim \mathcal{C}}{\mathbb{E}}\left\{ \frac{\mathrm{d} h_{C_i}\left(s_{t+q}\right)}{\mathrm{d} \theta}\right\} \\
=&\underset{s_{t} \sim \mathcal{C}}{\mathbb{E}}\left\{\sum_{j=t}^{t+q} \frac{\partial h_{C_i}\left(s_{t+q}\right)}{\partial s_{t+q}}\left[ \phi_{j-t}+ \psi_{j-t}\right]\right\}
\end{aligned}
\end{equation}
\begin{equation}
\begin{aligned}
&\text { where }\\
&\phi_{i+1}=\left\{\begin{array}{ll}
0 & , i=-1 \\
\frac{\partial f\left(s_{t+i}, a_{t+i}\right)}{\partial s_{t+i}} \phi_{i}+\frac{\partial f\left(s_{t+i}, a_{t+i}\right)}{\partial a_{t+i}} \psi_{i} & , \text { else }
\end{array}\right.\\
&\psi_{i+1}=\frac{\partial \pi\left(s_{t+i} ; \theta\right)}{\partial s_{t+i}} \phi_{i}+\frac{\partial \pi\left(s_{t+i} ; \theta\right)}{\partial \theta}
\end{aligned}
\notag
\end{equation}
According to Definition \ref{def:degree}, Each iterative item of $\psi_i$ is equal to zero, and $\frac{\mathrm{d} J_{C_i}}{\mathrm{~d} \theta}=0$. Therefore, if the rollout step is less than $m$, the input fails to affect constraints cost, and the constraints costs can not be optimized.
\end{proof}

\bibliographystyle{bib/IEEEtran}
\bibliography{bib/irosref}

\begin{thebibliography}{10}
\providecommand{\url}[1]{#1}
\csname url@samestyle\endcsname
\providecommand{\newblock}{\relax}
\providecommand{\bibinfo}[2]{#2}
\providecommand{\BIBentrySTDinterwordspacing}{\spaceskip=0pt\relax}
\providecommand{\BIBentryALTinterwordstretchfactor}{4}
\providecommand{\BIBentryALTinterwordspacing}{\spaceskip=\fontdimen2\font plus
\BIBentryALTinterwordstretchfactor\fontdimen3\font minus
  \fontdimen4\font\relax}
\providecommand{\BIBforeignlanguage}[2]{{%
\expandafter\ifx\csname l@#1\endcsname\relax
\typeout{** WARNING: IEEEtran.bst: No hyphenation pattern has been}%
\typeout{** loaded for the language `#1'. Using the pattern for}%
\typeout{** the default language instead.}%
\else
\language=\csname l@#1\endcsname
\fi
#2}}
\providecommand{\BIBdecl}{\relax}
\BIBdecl

\bibitem{RLBOOK}
\BIBentryALTinterwordspacing
S.~E. Li, \emph{{Reinforcement Learning and Control}}.\hskip 1em plus 0.5em
  minus 0.4em\relax Tsinghua University Lecture Notes, 2020. [Online].
  Available:
  \url{http://www.idlab-tsinghua.com/thulab/labweb/publications.html}
\BIBentrySTDinterwordspacing

\bibitem{amodei2016concrete}
D.~Amodei, C.~Olah, J.~Steinhardt, P.~Christiano, J.~Schulman, and
  D.~Man{\'{e}}, ``{Concrete problems in AI safety},'' \emph{arXiv preprint
  arXiv:1606.06565}, 2016.

\bibitem{Achiam2017a}
J.~Achiam, D.~Held, A.~Tamar, and P.~Abbeel, ``Constrained policy
  optimization,'' in \emph{Proceedings of the 34th International Conference on
  Machine Learning}, ser. Proceedings of Machine Learning Research, D.~Precup
  and Y.~W. Teh, Eds., vol.~70.\hskip 1em plus 0.5em minus 0.4em\relax
  International Convention Centre, Sydney, Australia: PMLR, 06--11 Aug 2017,
  pp. 22--31.

\bibitem{Chow2018a}
Y.~Chow, M.~Ghavamzadeh, L.~Janson, and M.~Pavone, ``{Risk-constrained
  reinforcement learning with percentile risk criteria},'' \emph{Journal of
  Machine Learning Research}, vol.~18, pp. 1--51, 2018.

\bibitem{Garcia2015a}
J.~Garc{\'{i}}a and F.~Fern{\'{a}}ndez, ``{A comprehensive survey on safe
  reinforcement learning},'' \emph{Journal of Machine Learning Research},
  vol.~16, pp. 1437--1480, 2015.

\bibitem{Uchibe2007}
E.~Uchibe and K.~Doya, ``Constrained reinforcement learning from intrinsic and
  extrinsic rewards,'' in \emph{2007 IEEE 6th International Conference on
  Development and Learning}.\hskip 1em plus 0.5em minus 0.4em\relax IEEE, 2007,
  pp. 163--168.

\bibitem{tessler2018reward}
C.~Tessler, D.~J. Mankowitz, and S.~Mannor, ``Reward constrained policy
  optimization,'' \emph{arXiv preprint arXiv:1805.11074}, 2018.

\bibitem{jorion2007value}
P.~Jorion, \emph{Value at risk: the new benchmark for managing financial
  risk}.\hskip 1em plus 0.5em minus 0.4em\relax The McGraw-Hill Companies,
  Inc., 2007.

\bibitem{Rockafellar2002}
R.~T. Rockafellar and S.~Uryasev, ``{Conditional value-at-risk for general loss
  distributions},'' \emph{Journal of Banking and Finance}, vol.~26, no.~7, pp.
  1443--1471, 2002.

\bibitem{Ray2019}
A.~Ray, J.~Achiam, and D.~Amodei, ``Benchmarking safe exploration in deep
  reinforcement learning,'' \emph{arXiv preprint arXiv:1910.01708}, 2019.

\bibitem{Duan2019b}
J.~Duan, Z.~Liu, S.~E. Li, Q.~Sun, Z.~Jia, and B.~Cheng, ``Deep adaptive
  dynamic programming for nonaffine nonlinear optimal control problem with
  state constraints,'' \emph{arXiv preprint arXiv:1911.11397}, 2019.

\bibitem{memarzadeh2019model}
M.~Memarzadeh and M.~Pozzi, ``Model-free reinforcement learning with
  model-based safe exploration: Optimizing adaptive recovery process of
  infrastructure systems,'' \emph{Structural Safety}, vol.~80, pp. 46--55,
  2019.

\bibitem{koller2018learning}
T.~Koller, F.~Berkenkamp, M.~Turchetta, and A.~Krause, ``Learning-based model
  predictive control for safe exploration,'' in \emph{2018 IEEE Conference on
  Decision and Control (CDC)}.\hskip 1em plus 0.5em minus 0.4em\relax IEEE,
  2018, pp. 6059--6066.

\bibitem{schulman2015trust}
J.~Schulman, S.~Levine, P.~Abbeel, M.~Jordan, and P.~Moritz, ``Trust region
  policy optimization,'' in \emph{International conference on machine
  learning}.\hskip 1em plus 0.5em minus 0.4em\relax PMLR, 2015, pp. 1889--1897.

\bibitem{Lin2021Solving}
Z.~{Lin}, J.~{Duan}, S.~E. {Li}, J.~{Li}, H.~{Ma}, Q.~{Sun}, J.~{Chen}, and
  B.~{Cheng}, ``Solving finite-horizon hjb for optimal control of
  continuous-time systems,'' in \emph{2021 International Conference on
  Computer, Control and Robotics (ICCCR)}, 2021, pp. 116--122.

\bibitem{prajna2006barrier}
S.~Prajna, ``{Barrier certificates for nonlinear model validation},''
  \emph{Automatica}, vol.~42, no.~1, pp. 117--126, 2006.

\bibitem{Agrawal2017a}
A.~Agrawal and K.~Sreenath, ``Discrete control barrier functions for
  safety-critical control of discrete systems with application to bipedal robot
  navigation.'' in \emph{Robotics: Science and Systems}, 2017.

\bibitem{Nguyen2016}
Q.~{Nguyen} and K.~{Sreenath}, ``Exponential control barrier functions for
  enforcing high relative-degree safety-critical constraints,'' in \emph{2016
  American Control Conference (ACC)}, 2016, pp. 322--328.

\bibitem{Ames2019a}
A.~D. {Ames}, S.~{Coogan}, M.~{Egerstedt}, G.~{Notomista}, K.~{Sreenath}, and
  P.~{Tabuada}, ``Control barrier functions: Theory and applications,'' in
  \emph{2019 18th European Control Conference (ECC)}, 2019, pp. 3420--3431.

\bibitem{ma2021feasibility}
H.~Ma, X.~Zhang, S.~E. Li, Z.~Lin, Y.~Lyu, and S.~Zheng, ``Feasibility
  enhancement of constrained receding horizon control using generalized control
  barrier function,'' \emph{arXiv preprint arXiv:2102.13304}, 2021.

\bibitem{Li2011}
S.~Li, K.~Li, R.~Rajamani, and J.~Wang, ``{Model predictive multi-objective
  vehicular adaptive cruise control},'' \emph{IEEE Transactions on Control
  Systems Technology}, vol.~19, no.~3, pp. 556--566, 2011.

\bibitem{Li2015}
S.~E. Li, Z.~Jia, K.~Li, and B.~Cheng, ``{Fast online computation of a model
  predictive controller and its application to fuel economy-oriented adaptive
  cruise control},'' \emph{IEEE Transactions on Intelligent Transportation
  Systems}, vol.~16, no.~3, pp. 1199--1209, 2015.

\bibitem{Guan2020Centralized}
Y.~{Guan}, Y.~{Ren}, S.~E. {Li}, Q.~{Sun}, L.~{Luo}, and K.~{Li}, ``Centralized
  cooperation for connected and automated vehicles at intersections by proximal
  policy optimization,'' \emph{IEEE Transactions on Vehicular Technology},
  vol.~69, no.~11, pp. 12\,597--12\,608, 2020.

\bibitem{Ren2020ITSC}
Y.~{Ren}, J.~{Duan}, S.~E. {Li}, Y.~{Guan}, and Q.~{Sun}, ``Improving
  generalization of reinforcement learning with minimax distributional soft
  actor-critic,'' in \emph{2020 IEEE 23rd International Conference on
  Intelligent Transportation Systems (ITSC)}, 2020, pp. 1--6.

\bibitem{ge2020numerically}
Q.~Ge, S.~E. Li, Q.~Sun, and S.~Zheng, ``Numerically stable dynamic bicycle
  model for discrete-time control,'' \emph{arXiv preprint arXiv:2011.09612},
  2020.

\end{thebibliography}

\end{document}